\documentclass{llncs}
\usepackage{subfigure,graphicx,url,times,multirow,amsmath,amssymb,algorithm,algorithmic,xspace,epsfig,todonotes,array,caption,hyperref}

\newcommand\tab[1][1cm]{\hspace*{#1}}
\newcommand{\keywords}[1]{\par\addvspace\baselineskip\noindent\keywordname\enspace\ignorespaces#1}

\usepackage[justification=centering]{caption}
\makeatletter
\renewcommand*\l@author[2]{}
\renewcommand*\l@title[2]{}
\makeatletter

\begin{document}
\mainmatter

\title{A Fast Evolutionary adaptation for MCTS in Pommerman}
\author{Harsh Panwar, Saswata Chatterjee, Wil Dube}
\institute{School of Electronic Engineering and Computer Science \\ Queen Mary University of London, UK \\ harshpanwar@ieee.org, s.chatterjee@se21.qmul.ac.uk, w.dube@se21.qmul.ac.uk }

\maketitle

\begin{abstract}

Artificial Intelligence, when amalgamated with games makes the ideal structure for research and advancing the field. Multi-agent games have multiple controls for each agent which generates huge amounts of data while increasing search complexity. Thus, we need advanced search methods to find a solution and create an artificially intelligent agent. In this paper, we propose our novel Evolutionary Monte Carlo Tree Search (FEMCTS) agent which borrows ideas from Evolutionary Algorthims (EA) and Monte Carlo Tree Search (MCTS) to play the game of Pommerman. It outperforms Rolling Horizon Evolutionary Algorithm (RHEA) significantly in high observability settings and performs almost as well as MCTS for most game seeds, outperforming it in some cases. 

\end{abstract}


\section{Introduction} \label{sec:intro}

Games are intuitive and fun which makes them rich in data generated by human-computer interaction which is very essential to do extensive research in AI. Experiments can be done to compare human intelligence with artificial intelligence (AI) by utilizing the 'fun' element which will attract users to play games without any particular incentive and generate more data to improve the AI. But not all games are fit for research. We need to look for a game which is fun to play for the users while still being intuitive and easy to implement. \\

\noindent
Pommerman \cite{resnick2018pommerman} is one such game with right amount of trade-offs. The benchmark is challenging due to the amount of hidden information it has making the element of intuition stronger. Due to the nature of multi-agent games the search tree formed by them is generally accompanied by a very large branching factor. At each step the branching factor (\emph{b}) can be as large as $6^4 = 1,296$ \cite{osogami2019real} , making it computationally expensive to use basic algorithms such as Breadth First search which have a run time complexity of $\mathcal{O}(b^d)$. On the other hand advanced search algorithms like MCTS and evolutionary algorithms are able to handle search trees with high branching factor as well. In search of even better results we have tried to combine MCTS with the genome-based method of evolutionary algorithms.  \\
\footnote{We have open sourced our code of proposed FEMCTS agent built on top of Pommerman implementation in Java by \cite{perez2019analysis}. \url{https://github.com/Neilchat/pommerman}.}

\noindent
The major contributions of this paper is:\\
\begin{itemize}
    \item We propose a novel Evolutionary Monte Carlo Tree Search agent in Pommerman. 
    \item The proposed agent is able perform almost as well as vanilla MCTS when seeded, and outperforms it in some cases. 
    \item The proposed agent is able to defeat RHEA and OSLA in most games. 
    \item We compare the performance of FEMCTS vs MCTS and RHEA in various observability settings. 
\end{itemize}

\noindent
Structure of the remaining paper is presented as Section \ref{sec:lit} focuses on the recent work done on AI in games, Pommerman, Evolutionary Algorithms and MCTS Section \ref{sec:bench} describes about the Pommerman framework that we have used as a benchmark for our algorithm Section \ref{sec:back} explains the basic background knowledge required to understand the proposed algorithm Section \ref{sec:meth} elaborate the proposed algorithms with detailed description \ref{sec:exp} reports about detailed experimental results along with the performance on Pommerman \ref{sec:discussion} and \ref{sec:conc} establishes the bridge between problem statement and the outcomes along with a discussion on future work. 

\section{Literature Survey} \label{sec:lit}

This section of our paper is dedicated to evaluating the previous work of other researchers on the enhancement of the MCTS algorithm particularly with evolutionary algorithms.\\

\noindent
\cite{Browne2012} survey an assortment of published work on MCTS and attempts to provide a comprehensive snapshot of its enhancement variations in the first five years of research on MCTS. \cite{Browne2012} contribute to the understanding the inner workings of MCTS and its family of variations, highlighting some of its strengths such as MCTS’ effectiveness in highly complex domains, and the fact that game play can be achieved with only just the rules with no knowledge of the game. Amongst these strengths is MCTS’ adaptability that allows it to be hybridised with other techniques, and even more importantly the fact that MCTS can still produce stronger play despite the noisy information from enhancement. Zeroing in on the given distribution data about the application of MTCS variations across a range of combinatorial games including Go, UCT (which is said to be MCTS with any UCB tree selection policy) appears to have been the most popular at the time, followed by Flat MC/UBC (said to be a Monte Carlo method with bandit-based move selection with no tree growth). \cite{Browne2012} also highlight MCTS’ weaknesses, which opens a new avenue of improvement challenges to researchers. MCTS was found to struggle to cope with an increase in the branching factor and depth of the graph. It was also found to be more effective as a hybrid in tandem with other techniques than on its own, meaning that it needs to be enhanced to produce plausible results, hence the continuing research effort over the years on various enhancements.\\

\noindent
\cite{swiechowski2021monte} review a few modifications and hybrid approaches, which is a similar work survey to the publication by \cite{Browne2012}. It's interesting to note that nearly a decade on, after seeing many hybrids of the standard version, MCTS’s shortcomings with regards to dealing tackling complex games with high branching factor are still being highlighted. \cite{swiechowski2021monte} also focus on modifications and extensions of the MCTS algorithm that reduces the complexity of a game  environment also citing the fact that the standard version of MCTS (formulated in 2006) is not good enough because its combinatorial complexity is high. \cite{swiechowski2021monte} establish that updating the playout policy parameters (particularly in deep neural networks), the gradient reinforcement learning method in the UCT variation of vanilla MCTS has seen an improvement in the strength of the algorithm. It also adds that including heavy playouts (as domain-specific knowledge) dramatically improves the performance of MCTS. It also reveals that using genetic algorithms results in a more effective and faster playouts function which was used to enhance MCTS policy in Pac-Man. While this \cite{swiechowski2021monte} do well to cover the improvements of MCTS by using domain knowledge, reinforcement and self-adaptive (evolution) techniques, its drawback is that it barely touches on when and where each of these enhancements thrive and fail. That is to say that the evaluation of techniques is not as extensive as in the comparative preceding survey from 2012.\\

\noindent
Having mentioned self-adaptive techniques earlier, \cite{sironi2018self} made a novel Self-Adaptive MCTS (SA-MCTS) agents that optimise the parameters of a non-Self-Adaptive MCTS agent of the GVGAI framework, whose results prove more robust. Much of this work feeds into the work of \cite{gaina2021rolling} who revisited the application of Rolling Horizontal Evolutionary Algorithm (RHEA) in games and used the N-Tuple Bandit Evolutionary Algorithm (NTBEA) to optimise the performance of RHEA, an experiment in which several parameters of the configuration of RHEA are modified, and as a result, the win rate of the agent over several games increased dramatically. Although \cite{sironi2018self} provide the evidence of some of the variations of RHEA outperforming MCTS in several games, it also investigates ways in which MCTS can be integrated into RHEA, the result of which doesn't prove conclusive. \\

 \noindent
\cite{baier2018evolutionary} assert that EMCTS is ineffectual for searching beyond the current player’s turn, which results in poor decision making, hence they extended EMCTS to search beyond the current turn using simple models and the action plan of the opponent. EMCTS works by combining the tree search of MCTS with Online Evolutionary Planning, a natural selection algorithm that is used for evolving and training. Flexible-Horizon EMCTS is said to expand the search horizon beyond the current turn, an attribute of MCTS that makes it ideal for complex adversarial games with multiple actions.\\

\noindent
Much of this project work builds around the work of \cite{lucas2014fast} on Fast Evolutionary Adaptation for MCTS is said to a new adaptive MCTS algorithm
that optimises its performance by using evolution and whose results are known to outperform MCTS in any case because it’s said to work with more informative statistics as a more diverse state space is explored by more decisive simulations, hence adaptation is almost immediate as the system learns. The method is detailed in the methodology, section 5. \\

\noindent
It is evident that there has been an improvement on MCTS since its inception in 2006. Several of its variants come into use in games over the years with the goal of improving the robustness and effectiveness of the algorithm and reducing the complexity of the environment traversed by the agent and also being able to contain a higher branching factor. It is needless to reiterate that these improvements are in line with the timely advent of more powerful computing systems.

\section{Pommerman} \label{sec:bench}

Pommerman \cite{resnick2018pommerman} is a multi-agent low dimensional four player game with discrete controls and communication channel amongst agent making it suitable for high level research on AI. Based on a classic game named Bomberman \cite{wikipedia_2021} developed by Hudson Soft in 1983, it is a recognised benchmark for multi-agent learning. Pommerman isn't the only competition running in this space, in fact, RoboCup \cite{kitano1997robocup} \cite{nardi2014robocup} has been used by AI researchers for a long time. Compared to RoboCup, Pommerman is easier to implement and experiment upon since the latter doesn't involve complexity arising due to robotics. One could argue that games like Counter-Strike which are fun and intuitive and doesn't have the sensors related difficulty would be better fitted for research in this field but unfortunately the implementation part makes Pommerman the better option. One single iteration (one full match) in Counter-Strike will result in a higher computational time complexity compared to one iteration in Pommerman since the controls in Counter-Strike are continuous. 

\begin{figure}
    \centering
    \includegraphics[scale=0.2]{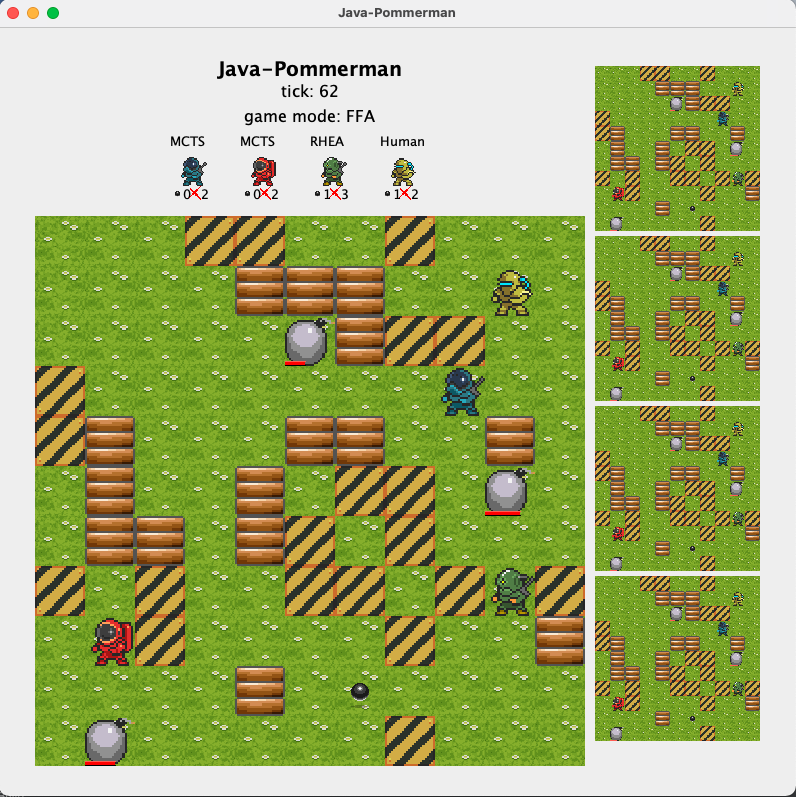}
    \caption{Snapshot of Pommerman implemented in Java}
    \label{fig:pommerman}
\end{figure}

\subsection{Game Rules}

\begin{itemize}
    \item The game is played on a 11x11 grid ('board') as seen in Fig.~\ref{fig:pommerman}.
    \item There are 4 agents in total and each player starts from one corner of the board.
    \item There are two type of walls:
    \begin{itemize}
        \item \textbf{Rigid walls}: An agent can't pass through rigid walls and can't destruct them.
        \item \textbf{Wooden walls}: A wooden wall can be destroyed by bombs. Once destroyed an agent can pass through them with or without a power up but not before that. 
    \end{itemize}
    \item An agent can take one out of six possible actions including moving up, right, left or down. It can also stop (counted as pass) or drop a bomb.
    \item Every agent starts uniformly with one bomb each which has the life of 10 tick after being dropped and a blast strength of 2 in all 4 directions. Once dropped the agent has to wait for the bomb to blast before it can plant a bomb again.
    \item Power Ups can be found in wooden walls after getting exploded. There are three type of power ups available:
    \begin{itemize}
        \item Extra bomb
        \item Increase range
        \item Kicking ability
    \end{itemize}
\end{itemize}

\noindent
There are various open-source implementations of Pommerman available including the Python implementation by \cite{DBLP} and Java implementation by \cite{perez2019analysis}. Due to being 45 times faster \cite{perez2019analysis} in processing time we have selected the Java implementation.

\section{Background} \label{sec:back}

\subsection{Monte Carlo Tree Search (MCTS)}
\begin{figure}
    \centering
    \includegraphics[scale=0.2]{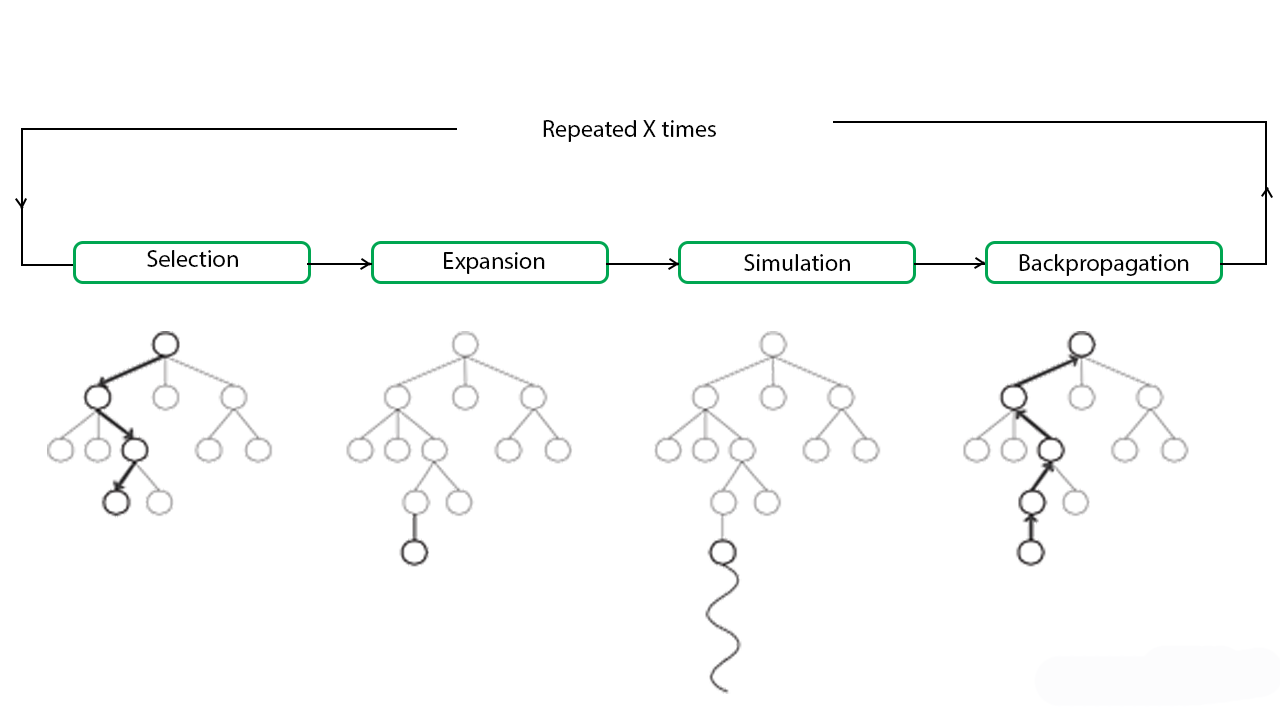}
    \caption{Four steps of MCTS}
    \label{fig:MCTS}
\end{figure}

MCTS is a search method which is highly selective, follows best-first search (BFS) and works well even with trees having high. An assymmetric tree is built which explores the most viable parts of the search space. The reason MCTS works so well is because it's based on random sampling action from states. Unlike in common terminology the word "randomness" holds a great value when used in statistics. \cite{kendall1938randomness} MCTS consists of four major steps (see Figure \ref{fig:MCTS}) :

\begin{itemize}

    \item \textbf{Selection:} Applied recursively until a leaf node is reached using a tree policy. 
    
    \item \textbf{Expansion:} New nodes are created.
    
    \item \textbf{Simulation:} One simulation of the game is performed.
    
    \item \textbf{Backpropogation:} The result of the simulation is backpropogated in the tree. 
    
\end{itemize}

\noindent
The significance of each node is that it represents a game state. And the following statistics are held in each node:

\begin{itemize}
    \item \textbf{\emph{N(s):}} Number of times the state \textbf{\emph{s}} has been visited by the algorithm.
    \item \textbf{\emph{N(s,a):}} Number of times the action \textbf{\emph{a}} has been played from state \emph{s}. 
    \item \textbf{\emph{Q(s,a):}} Approximation of how good it is to play action \textbf{\emph{a}} from \textbf{\emph{s}}
\end{itemize}

\subsection{Evolutionary Algorithms}

\noindent
Evolutionary algorithms is an optimization technique inspired by the  theory of evolution by natural selection [Charles Darwin, 1859]. A vanilla EA will generally consider these steps:

\noindent
\begin{itemize}
    \item \textbf{Initialization} A population in the form of an array is initialized randomly or by using some previous data. Each member in the population is known as sequence and each sequence has values known as genes. 
    \item \textbf{Evaluation} The population is then evaluated based on a fitness function. The function outputs a number for every sequence which is then maximised or minimised. 
    \item \textbf{Selection}  Randomly a sequence is selected from the population. The most common selection technique used is Tournament where \emph{M} sequences are selected from a population and the best one out of them is finalised.
    \item \textbf{Crossover} Two or more sequences are combined together to form a new (or more) sequence. 
    \item \textbf{Mutation} Mutation is a technique where randomly a gene (or more) inside a sequence is changed to bring new unique solutions in the population which may otherwise be never explored. 
\end{itemize}

\subsection{Evolutionary Monte Carlo Tree Search (eMCTS)}
Combination of the inspirations from theory of evolution with the Tree Search of MCTS results in eMCTS. Since both the algorithms rely heavily on randomness we can expect them to work together. The major additions in eMCTS are:
\begin{itemize}
    \item The node in every tree is replaced by a sequence of actions instead of a state.
    \item Leaf nodes are now evaluated by the fitness function instead of a simulation of the game. 
    \item A link from the parent to the child does not corresponds to a action but mutation. 
\end{itemize}

\subsection{Fast Evolutionary Adaptation for Monte Carlo Tree Search (FEMCTS)}
Another way of combining evolution and MCTS is to use evolution to tune hyper parameters for MCTS, i.e. the default policy and the tree policy. The idea is to maintain a population of individuals whose fitness is determined by the reward delta for an MCTS rollout and the individual in turn determines the Tree or/and the Default Policy. Evolving the population at the end of a sequence of rollouts would lead to an optimal population of individuals which determine tree and default policies to maximize reward. In this work we look at this variation of evolutionary MCTS in the context of Pommerman.

\section{Methodology} \label{sec:meth}

\subsection{Motivation}
The idea behind using evolutionary methods to enhance MCTS used here is to learn hyper-parameters of MCTS by maintaining a population that learns an optimal Default Policy. We implement this by determining the probability distribution over actions to use during the rollout that maximizes the reward at the end of it. Biasing rollouts can be very helpful when the search space is large and some paths are more useful than other and thus result in better recommendations if explored more.

\subsection{Unfeatured Fast Evolutionary MCTS}
We define an individual in our population as a set of weights. Each weight corresponds to the weight of an action in Pommerman. Using these weights we perform a biased rollout, the probability distribution over the actions obtained from the weights is the Default Policy. The probability distribution is computed using a softmax function over the action weights. For each root game state, the algorithm  iterates through the individuals in the population and perform MCTS rollouts using the Default Policy obtained from the individual's action weights. At the end of the rollout, the individual is evaluated using the reward value $\Delta$ found at the end of the rollout using a heuristic. After evaluating all the individuals in the population we evolve it; Selecting the best few (the number of elites is denoted by $E$ and here we used $E=4$) to stay on for the next generation and generating the others by performing tournament selection, uniform crossover and uniform mutation. Mutations are applied by adding or subtracting a small random value to the weight. To ensure we do not start from scratch at the beginning of each new game, we remember the population from the previous game and start of next game with it. To begin with, the population is initialized as arrays of zeros. The size of the population is denoted by $P$ and here we use $P=10$. Finally, the budget for the algorithm is defined by the stopping time and here we set it to 40ms.

\subsection{Featured Fast Evolutionary MCTS}
The strategy outlined above is not very smart as it tries to optimize the probability distribution for the rollouts without any knowledge of the state that the game is actually in. For example, it can't always be more beneficial to explore the branches where a bomb action is played, this would depend entirely on the game state. In order to get over this shortcoming, we consider individuals as weight matrices instead of arrays. Each individual's matrix $W$ has $nF$ columns and  $nEA$ rows. The algorithm works similar to the one described above. Only now, the Default Policy is calculated in a different way and depends on the game sate. For the game state at which the Default policy is to be calculated, first, a feature array $f$ of length $nF$ is calculated using the game state. Next, for each action $i$ we calculate the sum $\sum_{j} W_{i,j}f(j)$ over all the features. This gives us an array of weights for our actions, we perform a softmax over these weights to get the Default Policy. \\

\noindent
To evolve the population we again perform elitism, crossover and mutation. Selections are performed as usual via a tournament selection. Crossover is now performed for each action, each row of the offspring's weight matrix is constructed by performing a uniform crossover on the correspond row of the parents' matrices. To mutate, two strategies were explored. First, for each action a random set of features were chosen (configurable by the probability of mutation $mP$) and then the corresponding weights were mutated by either adding or subtracting a random number (configurable by mutation strength $mS$). In the second case, mutation was performed by choosing just one set of features for all the actions to mutate. These variations of mutation were explored in order to normalize the effect of mutations on the weights across features. The effect of both the types is quite similar and no significant differences in performances were observed between them, for the final set of experiments the second variation was used. The values of $mS$ explored were 0.01 (High), 0.001 (Medium) and 0.0001(Low). The high $mS$ proved to be too large for agent to perform well and was dropped from the final experimentation. $mP$ values explored were $0.5$ and $0.2$, the latter showed better performances and was used for the final experimentation.\\

\noindent
Feature selection was done by observing the game play to understand what might be important for the agent to understand to be better at the game. The complete feature array is of size 8 and contains bomb strength, whether the agent can kick,  the minimum distances to a bomb, a power up, an enemy, a rigid, a flame and a wood. The bomb strength was calculated as blast strength multiplied by ammunition divided by the 2 multiplied by the max bomb strength. The minimum distances were calculated using the Euclidean heuristic. These distances should be inversely related to the weight as their influence on the decision should be more when they are small, i.e the are nearby. Hence in the feature array they were calculated as $1/(distance +1)$. The addition of one is to avoid infinity values and bound this quantity from (0,1]. All the other features were normalized to have values less than or equal to one. An attempt was made at using Dijkstra to find better features from the game sate for example correct distances for all the different tile types etc, but this proved to be too computationally intensive. It brought down the number of iterations (one iteration being MCTS rollouts for each individual in the population) being performed by the agent by 2 folds.\\

\subsection{Improvements}
Next, we  clubbed actions together to get a set of effective actions for the game. The motivation behind this was the fact that the direction of movement was not really encoded in the features. If a bomb is near the agent, it should move, but its hard to say in which direction it should should move, that, can only be understood by rollouts. So, we now went from an action space of 6 to an effective action space of 3, move, stop and bomb ($nEA = 3$). The rest of the algorithm was kept as before. But now after calculating the action weights as $\sum_{j} W_{i,j}f(j)$, a factor of 4 was multiplied to the action weight corresponding to the move action before sending it to the softmax function. This is because the move effective action encodes 4 actual actions. For this to work, all the weights need to be positive, to achieve this we change the mutation function to mutate by adding a positive value 60\% of the time instead of 50\%. Another way to achieve this would be to start with a high positive value during initialization, this was not explored. If the move action was picked the algorithm would move randomly in any safe direction during the rollout. The other variation was one where the action space was just movement (including the stop action) or bomb ($nEA = 2$), in this case the action weight for movement is multiplied by 5. \\

\noindent
An improvement to this is to perform multiple rollouts per individual and then evaluate them with the average reward from the rollouts. After each individual is used for an MCTS run (selection, expansion and rollout) multiple times and they are evaluated, we evolve the population as described above. This is done so that the evaluation of each individual is better and gives a more robust final weight matrix. The number of rollouts we perform per individual is given by $L$ and we used a value of 4 for our experimentation.\\

\noindent
The last improvement was seeding the initial population. After playing the game 50 times with MCTS, RHEA and OSLA as opponents, a weight matrix where our agent was on a winning streak was taken and used as a seed for the next set of experiments.

\begin{algorithm}
\caption{Fast Evolutionary MCTS}
\label{Algo}

\textbf{Input:} $v_{o}$ \ root \ game \ state \\ 
\textbf{Output:} \ recommended \ action \ for \ root \ node  \ $v_{o}$ \\ \\ 
// initialize \ population \ \emph{(size, features, actions)} \\ 
1. \textbf{while} \ \emph{(within computational budget)} \\
2. \tab \textbf{for} \ \emph{(individual in population)} \\ 
3. \tab Initialize fitness Individual fitness stats \textbf{S} \\
4. \tab \tab \textbf{for} \ \emph{(i :=1 to K)} \\
5. \tab \tab \tab $v_{l}$ $\longleftarrow$ Tree Policy ($v_{o}$) \\
6. \tab \tab \tab $\Delta$ $\longleftarrow$ Default Policy (($v_{l}$), D(w)) \\
7. \tab \tab \tab Backup \ ($v_{l}$, $\Delta$) \\
8. \tab \tab \tab Update \ Individuals \ stats \ \textbf{S} $\longleftarrow$ $\Delta$ \\
9. \tab \tab Set \ Individual's \ fitness $\longleftarrow$ \ \textbf{S} \\
10. \tab evolve \ (population) \\
11. return \ a \  $\longleftarrow$ recommend ($v_{o}$) \\
\\
D(w) := $softmax(\sum{}{} w_{ij} f_{j})$ \\
$v_{o}$ := root node \\ 
$v_{l}$ := selected node \\
$L$ := Number of MCTS rollouts per individual fitness evaluation \\

\end{algorithm}

\section{Experimental Study} \label{sec:exp}

\subsection{Design}
A range of experiments were carried out on our algorithm to determine how well it performed in Pommerman. Our agent (FEMCTS) played 2 sets of 50 games against MCTS, RHEA and OSLA in which we used 10 game seeds and played 5 games per seed. We performed these runs for both the unfeatured algorithm and the two featured algorithms ($nEA = 3$ and $nEA = 2$). For the featured algorithms, we ran some simulations with the same configurations but included an initial seeding for the population, selecting the seed from a winning streak on the previous unseeded run.

\subsection{Unfeatured Agent}

Average of the two 50 game runs:
\\ \\
\begin{center}
\begin{tabular}{ |c|c|c|c|c| } 
 \hline
  	     Win &	Tie & Loss & Player \\ 
  	     \hline
  	26\%	&16\%	&58\% & EMCTS  \\
    0\% &	1\%	&99\% & OSLA \\
    22\% &	12\%	&66\% & RHEA  \\
  	34\%	& 18\%	&48\% & MCTS  \\
 \hline
\end{tabular}
\captionof{table}{Performance for unfeatured agent}\label{sophisticatedtable}
\end{center} 
 \vspace{2mm}
 
 \noindent
 The low performance in this result is expected as the unfeatured algorthim searches the tree without any information of the game state, biasing rollouts blindly.

\subsection{Featured Agent ($nEA = 3$)}

Here we have an effective action space move, stop and bomb. Feature space of size 8 containing bomb strength, whether the agent can kick,  the minimum distances to a bomb, a power up, an enemy, a rigid, a flame and a wood. The average of the two 50 game runs and the effect of adding an initial seeding with low mutation rate is as follows. 
\\ \\
\begin{center}
\begin{tabular}{ |c|c|c|c|c|c|c|c|c|c| } 
 \hline
 & Unseeded  & & & & Seeded (L)& &\\
 \hline
  	 & Win &	Tie & Loss & & Win &	Tie & Loss \\ 
  	 \hline
  FEMCTS	&18\%	&16\%	&66\% & 	&20\%	&26\%	&54\% \\
  OSLA &0\% &	1\%	&99 &  &0\% &	2\%	&98 \\
  RHEA &25\% &	14\%	&62\% &  &18\% &	20\%	&62\% \\
  MCTS&40\%	& 17\%	&43\% & &34\%	& 22\%	&44\% \\
 \hline
\end{tabular}
\captionof{table}{Performance of seeded and unseeded 3 effective action agent }\label{sophisticatedtable}
\end{center} 

\vspace{3mm}

\noindent
We can see that the seeded agent did slightly better than the unseeded one, although it performed worse than the unfeatured agent. This is probably down to randomness and 50 games were not enough to tell them apart.

\subsection{Featured Agent ($nEA = 2$)}

Here we have an effective action space of movement and bomb. Feature space of size 4 containing the minimum distance for bombs, enemies, power ups, rigid walls and whether the agent can kick. The average of the unseeded runs is given below, along with adding an initial seeding with two levels of mutation low (Seeded (L)) and high (Seeded (M) . \\ \\

\noindent
\begin{center}
\begin{tabular}{ |c|c|c|c|c|c|c|c|c|c|c|c|c| } 
 \hline
 &Unseeded  & & & & Seeded (L)& & & &Seeded (M) & & \\
 \hline
  	        & Win &	Tie & Loss & & Win &	Tie & Loss & & Win &	Tie & Loss \\ 
  \hline
  FEMCTS	   &23\%	&23\%	&49\% &     &30\%	&24\%	&46\% & 	&22\%	&40\%	&38\% \\
  OSLA     &1\%     &0\%	&99\% &     &0\%    &0\%	&100  &     &0\%    &0\%	&100 \\
  RHEA     &18\%    &23\%	&59\% &     &18\%   &26\%	&56\% &     &16\%   &28\%	&56\% \\
  MCTS     &26\%	& 26\%	&48\% &    &22\%	&30\%	&48\% &     &20\%	& 42\%	&38\% \\
 \hline
\end{tabular}
\captionof{table}{Performance of the three versions 2 effective action agent in FFA mode. }\label{sophisticatedtable}
\end{center}

\vspace{3mm}

\noindent
As this looked like a promising agent and we decided to run a large 200 game simulation with low mutation, the result was as follows:

\vspace{3mm}
\noindent
\begin{center}
\begin{tabular}{ |c|c|c|c|c| } 
 \hline
  N	    & Win &	Tie & Loss & Player \\ 
  \hline
  200	&26\%	&20\%	&54\% & FEMCTS \\
  200  &1\% &	1\%	&98 & OSLA\\
  200  &18.5\% &	20.5\%	&61\% & RHEA\\
  200	&32\%	& 21\%	&47\% & MCTS\\
 \hline
\end{tabular}
\captionof{table}{Performance for seeded 2 effective action agent in FFA mode for 200 games}\label{sophisticatedtable}
\end{center} 

\vspace{3mm}

\noindent
This result shows that the agent could do well in some games but not consistently enough. This might be down to the game seed or the number of simulations we are performing.
\vspace{5mm}

\noindent
Finally, we chose the variant that performed the best in the above simulations, i.e. the featured $nEA = 2$, seeded agent and performed some experiments with playing in team mode and with partial observability. 

\subsection{Team Mode}

In team mode with full observability we ran two simulations. We played a team of our agents vs a team of MCTS agents and a team of RHEA agents. The simulation was for 10 seeds and 5 games played per seed. 

\vspace{3mm}
\noindent
\begin{center}
\begin{tabular}{ |c|c|c|c|c| } 
 \hline
  Simulation & Win &	Tie & Loss & Player \\ 
  \hline
  VS MCTS & 28\%	&12\%	&60\% & FEMCTS \\
    & 60\%	& 12\%	&28\% & MCTS\\
 \hline
  VS RHEA & 66\%	&6\%	&28\% & FEMCTS \\
    & 28\%	& 6\%	&66\% & RHEA\\
 \hline
\end{tabular}
\captionof{table}{Performance for FEMCTS team vs MCTS team and RHEA team}\label{sophisticatedtable}
\end{center} 
\noindent
In team mode with full observability we see expected results where our agent's team outperforms RHEA but is outperformed by MCTS.

\subsection{Partial Observability}

\noindent
We performed a simulation of 50 game runs against a team of two RHEA and two MCTS opponents with observability of 0, 1, 2, 3 and 4. The results for our agent's team were as follows: \\ \\
\vspace{3mm}
\begin{center}
\begin{tabular}{ |c|c|c|c|c|c| } 
 \hline
 & RHEA & & &MCTS & \\
 \hline
  PO   & Win &	Tie & & Win &	Tie  \\ 
  \hline
  0	&31\%	&0\%&	 &0\%	&0\%\\
  1	&40\%	& 0\%&   &6\%	& 0\%\\
  2	&46\%	& 0\%&   &8\%	& 20\%\\
  3	&45\%	& 5\%&   &38\%	& 20\%\\
  4	&64\%	& 2\%&   &36\%	& 20\%\\
 \hline
\end{tabular}
\captionof{table}{Performance of a Team of FEMCTS agents vs team of RHEA agents}\label{sophisticatedtable}
\end{center}

\vspace{3mm}

\noindent
Clearly our agent performs better with increasing observability, outperforming RHEA when observability is at 4, as expected it is outperformed by MCTS.\\ \\
In FFA mode with partial observability we played our agent against MCTS, RHEA and OSLA like before. The following is a  report for the win percentages for all the observability settings: \\ \\
\noindent
\begin{center}
\begin{tabular}{ |c|c|c|c|c| } 
 \hline
  PO   & MCTS &	FEMCTS & RHEA & OSLA  \\ 
  \hline
  0	&94\%	&0\% & 4\% & 0\%	 \\
  1	&66\%	& 10\% & 22\% & 0\%\\
  2	&54\%	& 10\% & 28\% & 0\%\\
  3	&40\%	& 22\% & 24\% & 0\%\\
  4	&32\%	& 28\% & 20\% & 0\%\\
 \hline
\end{tabular}
\captionof{table}{Win percentages in FFA mode on various observability settings}\label{sophisticatedtable}
\end{center}

\begin{figure}
    \centering
    \includegraphics[scale=0.5]{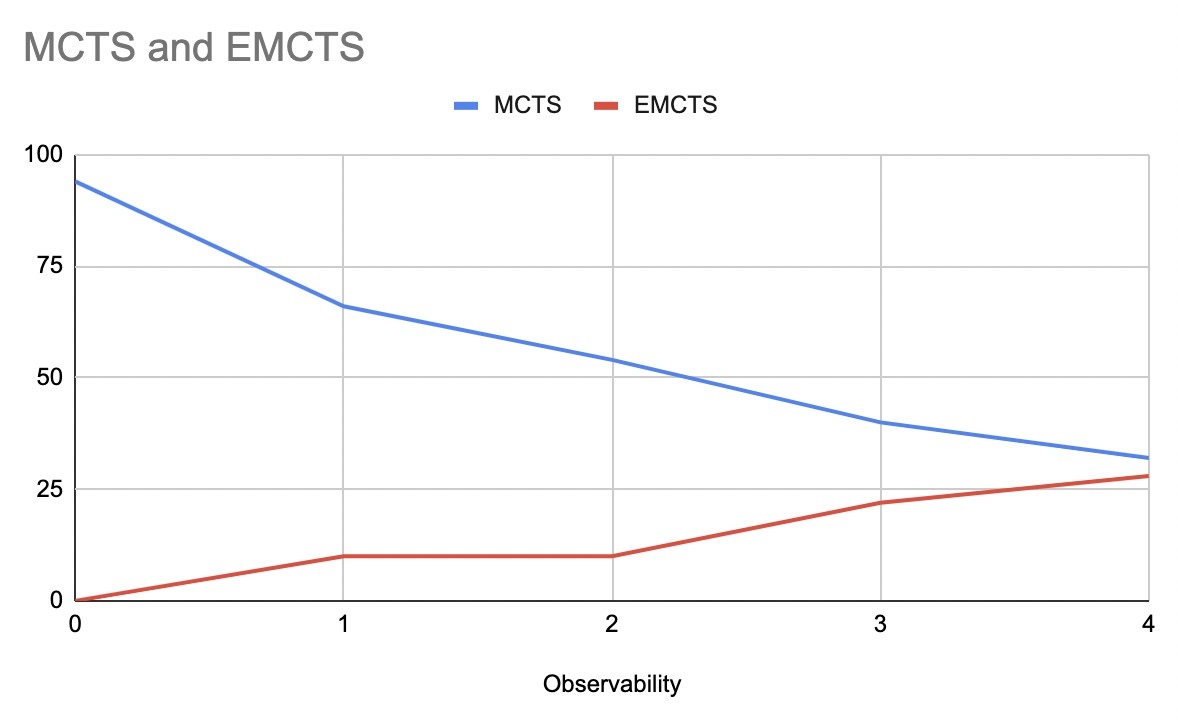}
    \caption{Performance (win percentages) of MCTS vs FEMCTS in various observability settings for 50 games played in FFA mode. We can see the performance of EFMCTS increase with increase in observability, almost matching MCTS at observability = 4.}
    \label{fig:pommerman}
\end{figure}
\vspace{5mm}
\begin{figure}
    \centering
    \includegraphics[scale=0.5]{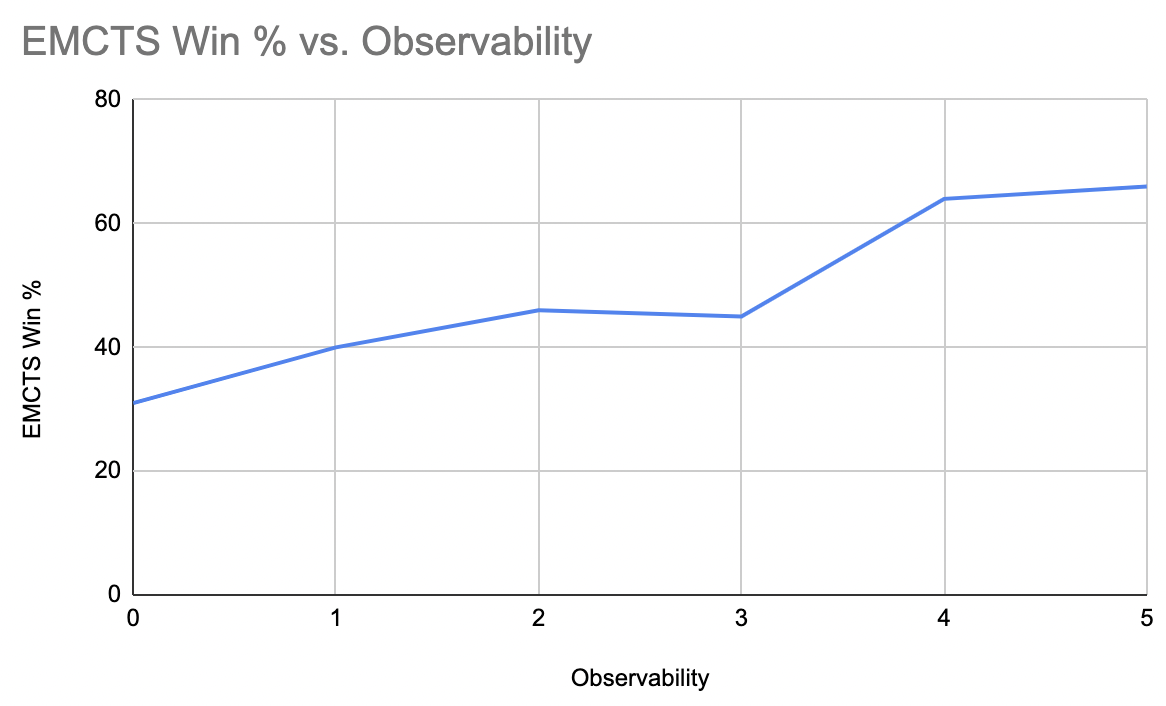}
    \caption{Win percentages of FEMCTS team vs RHEA team in various observability settings for 50 games played in Team mode. Here observability = 5 is full observability. Again, an increase in performance with observability is observed, finally outperforming the RHEA team at observabity = 4.}
    \label{fig:pommerman}
\end{figure}

\vspace{50mm}

\section{Discussion} \label{sec:discussion}

\noindent
Our agents were consistently better than RHEA but not against MCTS.
Convergence for the weight matrices to an optimal proved difficult in all the versions of our algorithm. The ordering of the weights kept changing. In this case we were concerned only about the internal ordering of each column, i.e. how much a particular feature forces a certain action to be explored. The reason for this could be that the features used are not great to predict more promising areas of the tree. Our features are not effective when there is nothing in the proximity of the agent. Also, they use a heuristic distance instead of a proper path distance as Dijkstra proved to be too computationally expensive. The features are also unweighted themselves, and perhaps having a weight for each of them could help in convergence as some features could be more important than others. Furthermore, there were features that were clearly more predictive than others. The action weights for the features' minimum distance to a bomb or an enemy often ended up in the same ordering, especially when the agent was winning. It is more helpful to move if there was a bomb, and more helpful to place a bomb if there was an enemy nearby. When the agent was using too many features, this ordering appeared to be absent more often.\\

\noindent
For the two-action variant, the results were closely matching those of vanilla MCTS on several runs. This is because when we are biasing just movement versus bomb, we really aren't biasing much. For the three-action, we are doing a little more than this, as we are biasing three action types. Finally, in the all actions version we are doing more than what the features have any knowledge about. It was observed that the weight matrix at tick 500, for the game that our agent won, often appeared to be the one that made sense for the game. This suggests that seeding the agent would help, which proved to be true. In general, the wins rate increased when we seeded and when we reduced the mutation strength. This indicates that evolution is struggling to find an optimal matrix and hold to it. This can be due to the tuning of the parameters such as the population size, elite count, whether to mutate elites or not, mutation strength and mutation probability. Another factor that influences this is the number of evaluations per individual $L$ used to compute its fitness. \\

\noindent
During game play the agent was observed to commit suicidal moves at times, when it played a bomb and moved to a position from which it’s trapped and can't move. This is counterintuitive because after planting the bomb, a move into a dead end gets a bad reward and should be strongly avoided. It could be that the agent performs well in some game seeds and not so well in others, where it traps itself and loses the game. This could explain why our best agent is observed doing much better in some smaller runs (10 seeds X 5 games each) but ultimately not so well at larger runs (20 seeds X 10 games each).\\

\noindent
Finally, partial observability appears to to be hard problem for our agent to tackle, especially in low observability settings. The FEMCTS team doesn't survive against the MCTS team even until game tick 500 at zero observability. We see that with increasing observability the FEMCTS team performs better, finally performing nearly as well as the MCTS team at full observability. In the experiments conducted against the RHEA team, our agent's team performs better with increasing observability and finally outperforms the RHEA team at observability = 4. On average the games where the FEMCTS team does better lasts longer. A sensible justification for this observed behaviour could be the fact that our agent uses features of the game state to inform search directions. It doesn't gain much from the features at low observability. Subsequently, it ends up exploring the wrong branches more often than it should be doing and hence choosing the action for the root node. It might prove beneficial to unseeded them for low observability settings.

\section{Conclusions and Future Work} \label{sec:conc}

The implementation of the evolution to determine the best Default Policy for MCTS for Pommerman could have been a success had the solution converged. It appears to be a promising area that can be explored in greater depth because the agent outperformed MCTS on several occasions. On these occasions the weight matrix evolved to a possible optimum to give the agent a winning streak. The issue of convergence needs to be addressed. It might be that this is not at all possible in Pommerman since there is no optimal weight matrix for all game seeds. It is more likely that this is to do with the interpretation of the game state. This means that our feature space needs to be improved in order to converge upon an optimal weight matrix.\\

\noindent
A possible modification to the agent that could be explored further would be to normalize the weight matrices in some fashion. This might help with convergence, as at each step we are concerned more with ratios and ordering between the weights rather than the actual values. Another improvement might be to weight the reward $\Delta$ added to the nodes during backpropagation by multiplying them by the factor of $\lambda^k$, where $\lambda$ is a value between $0$ and $1$, and $k$ is the depth. This would ensure that the reward coming from near the top of the tree is valued more as the nodes are more frequently visited. Some experiments could also be done to localize the issues the agent faces. We could try to identify game seeds on which the agent performs better if any, and use it to understand why it does so.

\pagebreak
\bibliographystyle{apalike}
\bibliography{Biblio}

\end{document}